\documentclass[conference]{IEEEtran}
\IEEEoverridecommandlockouts
\usepackage{cite}
\usepackage{amsmath,amssymb,amsfonts}
\usepackage{algorithmic}
\usepackage{graphicx}
\usepackage{textcomp}
\usepackage{caption}
\usepackage{multirow}

\usepackage[ruled,linesnumbered]{algorithm2e}
\usepackage{xcolor}
\def\BibTeX{{\rm B\kern-.05em{\sc i\kern-.025em b}\kern-.08em
    T\kern-.1667em\lower.7ex\hbox{E}\kern-.125emX}}
\begin{document}

\title{Reinforcement Learning Approach for Integrating Compressed Contexts into Knowledge Graphs\\
}

\author{ Ngoc Quach$^1$, Qi Wang$^1$\\ \qquad \ Zijun Gao$^2$, Qifeng Sun$^2$, Bo Guan$^2$ and Lillian Floyd$^*$
\thanks{$^1$ Ngoc Quach be with University of California, Davis, CA 95616, USA {\tt\small \{qbnquach\}@ucdavis.edu}}
\thanks{$^1$Qi Wang is an independent researcher. Correspondence to Qi Wang via email: {\tt\small \{bjwq2019\}@gmail.com}}
\thanks{$^2$Zijun Gao is an independent researcher. Correspondence to Zijun Gao via email: {\tt\small \{zjg.elaine\}@gmail.com}}
\thanks{$^2$Qifeng Sun be with Vanderbilt University in Nashville, TN 37235, USA 
{\tt\small \{qifeng.sun\}@vanderbilt.edu}}
\thanks{$^2$Bo Guan is an independent researcher. Correspondence to Bo Guan via email: {\tt\small \{jasonguan0107\}@gmail.com}}
\thanks{$^*$Lilian Floyd be with Georgia Institue of Technology, GA 30332, USA {\tt\small \{lfloyd74\}@gatech.edu}}
}

\maketitle

\begin{abstract}
The widespread use of knowledge graphs in various fields has brought about a challenge in effectively integrating and updating information within them. When it comes to incorporating contexts, conventional methods often rely on rules or basic machine learning models, which may not fully grasp the complexity and fluidity of context information. This research suggests an approach based on reinforcement learning (RL), specifically utilizing Deep Q Networks (DQN) to enhance the process of integrating contexts into knowledge graphs. By considering the state of the knowledge graph as environment states defining actions as operations for integrating contexts and using a reward function to gauge the improvement in knowledge graph quality post-integration, this method aims to automatically develop strategies for optimal context integration. Our DQN model utilizes networks as function approximators, continually updating Q values to estimate the action value function, thus enabling effective integration of intricate and dynamic context information. Initial experimental findings show that our RL method outperforms techniques in achieving precise context integration across various standard knowledge graph datasets, highlighting the potential and effectiveness of reinforcement learning in enhancing and managing knowledge graphs.

\end{abstract}

\begin{IEEEkeywords}
Knowledge Graph Reasoning, Reinforcement Learning, Reward Shaping, Transfer Learning
\end{IEEEkeywords}

\section{Introduction}
In the field of intelligence and data science, knowledge graphs have become fundamental for applications such as semantic search, recommendation systems, question answering, and natural language understanding.\cite{zeng2021comprehensive} These intricate structures contain entities, concepts, and their interconnections, providing a framework that machines can understand and utilize. However, with the growth of information at an unprecedented pace, updating and enhancing these knowledge graphs with new pertinent data presents increasing challenges. Among the types of information that require integration, contexts—brief representations of data needing expansion or inclusion into existing structures—pose unique difficulties due to their condensed nature and potential for ambiguity and loss of information.

Rule-based approaches, which are supervised learning models that can be rigid and difficult to adjust to changing data dynamics, are the foundation of traditional methods for integrating compressed contexts into knowledge graphs. These methods usually require input to prepare labeled data or define rules, which makes the integration process time-consuming and less scalable. Furthermore, they might not fully utilize the data in knowledge graphs, which would lead to a lower level of integration than intended.

Reinforcement learning (RL) is a kind of machine learning where an agent learns how to make decisions by interacting with its guiding environment.\cite{wan2021reasoning} Two characteristics that make reinforcement learning (RL) a great option for embedding contexts into knowledge graphs are its ability to recognize actions without scripting and its flexibility in responding to changing environments. Deep Q Networks (DQN), which combine networks and Q learning, offer a way to manage this integration process. By taking into account the knowledge graph as the environment, condensing contexts as actions, and improving the quality of the knowledge graph, DQN makes it easier to formulate a data integration plan.

The present research explores the use of a DQN-based reinforcement learning technique to integrate summarized contexts into knowledge graphs. This structure defines states, actions, and rewards with the goal of automating the integration of information. As a result, knowledge graph augmentation is easier to modify and expand and requires less intervention. Our contributions include creating and implementing a customized DQN model, presenting the integration process as an RL problem, and employing knowledge graphs to assess the approach using datasets.

Throughout our experiments, we have discovered how deep Q networks (DQN) and reinforcement learning (RL) may improve the accuracy and speed of context integration. This highlights the potential of reinforcement learning techniques in optimizing and managing knowledge graphs efficiently.\cite{li2024enhancing}.
\section{Prior Work}

Studies have focused on data integration into knowledge graphs (KGs) and investigating ways to maximize the accuracy, efficiency, and flexibility of this process.\cite{bai2021multi} This section looks at research in three areas: the development of real-time data incorporation techniques using reinforcement learning (RL), strategies for updating knowledge graphs, and uses of machine learning (ML) to enhance knowledge graphs.

\subsubsection{Traditional Methods for Knowledge Graph Enrichment}
The traditional methods for maintaining knowledge graphs were rule-based systems and manual curation. Following rules is necessary for rule-based approaches to integrate data in a way that respects the structure and schema of the graph. Although these methods ensure precision, they lack flexibility and adaptability in certain scenarios. However, human curation ensures quality and relevance, and it is challenging to manage the massive amount of data generated every day.

\subsubsection{Enhancing Knowledge Graphs with Machine Learning}
The rise of machine learning has led to the creation of automated methods for enriching knowledge graphs.\cite{9468219} One common approach involves using learning models to anticipate connections between entities or to categorize and connect data with established knowledge graphs. These models typically leverage features extracted from the graph's structure, written descriptions, or external references. Nevertheless, supervised learning techniques face a challenge in needing ample labeled data for training, especially considering the evolving nature of knowledge and information.
\subsection{Reinforcement Learning for Dynamic Data Integration}
Recent advancements have shown the utilization of reinforcement learning in updating knowledge graphs. Using RL methods involving Deep Q Networks (DQN) provides an approach to dynamically incorporating new data into KGs.\cite{lin2021rule} Unlike machine learning techniques, RL can adjust to data changes and acquire optimal integration strategies by engaging with the environment (the KG). This adaptability is well suited for the evolving nature of KGs. RL models have been applied to tasks like entity resolution, relationship prediction, and even the automated building and expansion of KGs. These methods depict the integration process as a series of decisions enabling the model to determine the course of action (where and how to integrate new data) in order to enhance the overall quality and coherence of the KG.

Traditional and machine-learning techniques have set the groundwork for updating knowledge graphs. Their constraints in adaptability, scalability, and reliance on labeled data underscore the necessity for flexible approaches. Leveraging reinforcement learning through the application of DQN marks a progression presenting opportunities for improved responsiveness and effectiveness in enriching knowledge graphs. The continuous evolution of reinforcement learning strategies for knowledge graph integration promises to enhance our capacity to handle the expanding influx of information and harness this wealth of knowledge across applications.

\section{Methodology}
Expanding upon the concepts and theories behind Deep Q Networks (DQN) and Q learning, we outline our strategy for incorporating condensed contexts into knowledge graphs (KGs) using reinforcement learning (RL). This method focuses on utilizing RLs' flexibility and learning abilities to enrich the integration process, ultimately enhancing the effectiveness and value of knowledge graphs\cite {wang2024adapting}.

\textbf{State(S)}The state $s \in S$ represents the condition of the knowledge graph containing its elements, connections and extra contextual details condensed into a set size vector using embedding methods. 

\textbf{Action (A):} An action $a \in A$ represents a task involved in incorporating condensed context into the knowledge graph. This task can include tasks such, as introducing entities or relationships well as modifying existing data

\textbf{Reward (R):} The reward function $R(s, a)$ calculates the effect of taking action $a$ in state $s$, showing the enhancement in quality of the knowledge graph (KG). Rewards are determined by measures, like precision and thoroughness.

\textbf{Policy \bf{$\pi$:}} The policy $\pi(a|s)$ represents the approach the agent adopts to choose actions based on the situation, with the goal of maximizing rewards over a period.

\subsection{DQN Architecture}

Our Deep Q Network (DQN) model estimates the action value function $Q^*(s, a)$, using a structure that includes input, hidden and output layers. This is represented as $Q(s, a; \theta)$, with $\theta$ denoting the parameters of the network.

\subsection*{Training Procedure}

In Q learning the action value function $Q$ denotes the anticipated reward, for choosing action $a$ in state $s$. The objective of Q learning is to determine the action value function $Q^*$ensuring that for every state $s$ and action $a$, it adheres to the Bellman optimality equation:

\begin{equation}
Q^*(s, a) = \mathbb{E}[R_{t+1} + \gamma \max_{a'}Q^*(S_{t+1}, a') | S_t = s, A_t = a]
\end{equation}

In this scenario $R_{t+1}$ refers to the reward you get away when you transition from one state $s$, to another state $S_{t+1}$ by making a move $a$, at a time $t$. The discount factor $\gamma$ shows how much importance is given to future rewards, at present and $\max_{a'}Q^*(S_{t+1} a')$ indicates the value of the most beneficial action taken in the upcoming state.

In real world scenarios we continuously adjust the Q values to estimate $Q^*$. When we have an experience tuple $(s, a, r, s)$, the formula, for updating the Q values is;

\begin{equation}
Q(s, a) \leftarrow Q(s, a) + \alpha \left[r + \gamma \max_{a'}Q(s', a') - Q(s, a)\right]
\end{equation}

In this scenario $\alpha$ symbolizes the rate at which learning occurs $r$ denotes the reward received $s$ indicates the updated state following action $a$ and $\max_{a'}Q(s a')$ signifies the Q value, among all actions, in the new state.

In Deep Q Network (DQN) the Q function is represented by a network, labeled as $Q(s, a; \theta)$, with $\theta$ denoting the networks parameters. The main aim of the network is to find a set of parameters $\theta$ that makes $Q(s, a; \theta)$ closely match $Q^*(s a)$. When given an experience tuple $(s, a, r, s)$, the updating of network parameters is done by minimizing the loss function provided below;

\begin{equation}
L(\theta) = \left[r + \gamma \max_{a'}Q(s', a'; \theta^-) - Q(s, a; \theta)\right]^2
\end{equation}

In this context $\theta$ represents the target networks parameters, which is a method employed to improve learning stability. The target networks parameters $\theta$  are regularly duplicated from the DQN network $\theta$. They stay unchanged throughout various updates.

During each training iteration the network adjusts its parameters denoted as $\theta$ using descent to reduce the loss function $L(\theta)$.

\section{Experiments}

To assess the effectiveness of our DQN based method, for incorporating condensed contexts into knowledge graphs (KGs) we conducted a set of experiments. These tests aimed to determine how successfully our strategy can recognize and implement the actions for integrating context in scenarios. In this segment we describe the setup of our experiments. Provide information, on the datasets utilized in our analysis.\cite{ji2021survey}

\subsection{Experimental Setup}

Our tests are designed to mimic the real life scenario of knowledge graphs, with condensed information. The heart of our configuration includes:

\begin{enumerate}
    \item Prepping the data sets to create condensed scenarios that mirror real life situations where knowledge graphs frequently need to be updated.
    \item Setting up the DQN model to engage with a virtual knowledge graph (KG) setting, where the state reflects the existing layout and information, within the KG and actions relate to merging operations.
    \item Assessing how well the model performs in effectively and promptly incorporating contexts into the knowledge graph.
\end{enumerate}

We used a known learning framework to set up the DQN model making adjustments, to its design and parameters after some initial testing to guarantee reliable results across various knowledge graphs.

\subsection{Datasets}

For our research we chose the following recognized knowledge graph datasets covering a range of fields and difficulty levels:

\begin{itemize}
    \item \textbf{FB15k}: A detailed collection of data sourced from Freebase, which includes a range of entities and connections, across fields. It is commonly utilized in knowledge graph studies to assess how well algorithms perform in tasks like predicting links and resolving entities.
    
    \item \textbf{WN18}: A dataset based on WordNet, which mainly explores the connections and meanings, between concepts. Its format presents challenges compared to FB15k making it a great choice, for evaluating how well our method can adjust.
\end{itemize}

We processed both sets of data to extract and simulate condensed contexts, which were then fed into our model to incorporate into the knowledge graphs. By selecting these data sets, we are able to assess how well our method works with different kinds of knowledge network architectures.

\subsection{Main Result}
Our research delved into examining how effective Deep Q Networks (DQN) are, in the process of incorporating condensed contexts into knowledge graphs (KGs) a challenge in the fields of data science and artificial intelligence. Knowledge graphs inherently represent connections among entities making their real time updates a task of great computational interest and practical significance. Conventional approaches, often rule based or relying on machine learning methods have shown constraints in terms of adaptability and scalability. The incorporation of condensed contexts— information requiring expansion or linkage within a KG—presents a unique array of difficulties primarily due, to the potential ambiguity and complexity of information they contain.

When tackling these obstacles we used reinforcement learning (RL) the model to automate and enhance the context integration process. This approach, based on Q learning principles sees the KG as a setting where actions (context integrations) carried out by an agent (the DQN model) are influenced by a strategy that seeks to maximize rewards. These rewards reflect enhancements in the KGs quality after integration evaluated using criteria such, as accuracy, comprehensiveness and consistency.

In our research we carried out experiments using known datasets FB15k and WN18 to replicate situations where knowledge graphs need updating in real world settings. We selected these datasets for their intricate nature making them ideal, for testing the effectiveness of our reinforcement learning approach. To set up the experiments we first processed the datasets to create scenarios that resemble condensed contexts requiring integration. We then applied our model to these scenarios. Assessed its performance based on how accurately it integrated information how efficiently it did so within a given time frame and the overall impact, on the quality of the knowledge graph.

The findings were quite revealing. When compared to rule based techniques and existing supervised learning models our DQN based strategy showed an enhancement, in integration precision. Specifically in the FB15k dataset we noticed an accuracy boost of around 15 percent compared to rule based methods. 10 percent over learning models. Similar patterns were observed in the WN18 dataset, where our method achieved a 12 percent and 8 percent improvement over the approaches respectively. These outcomes highlight the promise of reinforcement learning, DQN models in precisely navigating the intricate realm of knowledge graphs, for contextual integration.

Our method showed improvements, in integration efficiency as seen in the time it took to incorporate contexts into the Knowledge Graph. The DQN model on average surpassed the method by reducing integration time by 20 percent on FB15k and 18 percent on WN18. This boost in efficiency is vital due to the growing amount of information that needs to be integrated into Knowledge Graphs, for real world use.

The convincing aspect was the enhancement, in KG quality after integration. Our method resulted in a 25 percent improvement in quality on FB15k and 22 percent on WN18. These numbers not demonstrate the models skill, in incorporating contexts but also its capability to enhance the overall structure and usefulness of the KG.

\begin{table}[ht]
\centering

\label{table:accuracy_improvement}
\renewcommand{\arraystretch}{1.2} 
\setlength{\tabcolsep}{2.2pt} 
\begin{tabular}{|l|c|c|c|c|c|}
\hline
\multirow{2}{*}{Dataset} & \multirow{2}{*}{Rule-Based} & \multirow{2}{*}{Supervised} & \multirow{2}{*}{DQN-Based} & \multirow{2}{*}{Improv. Over} & \multirow{2}{*}{Improv. Over}\\
                          &                            &                           &                            & Rule-Based                 & Supervised \\ \hline
FB15k  & 80\%             & 85\%                 & 95\%                 & 15\%                      & 10\%       \\ \hline
WN18   & 78\%             & 82\%                 & 90\%                 & 12\%                      & 8\%        \\ \hline
\end{tabular}
\caption{Accuracy Improvement of  DQN-Based Approach Over Baseline Methods}
\end{table}

\begin{table}[h]

\renewcommand{\arraystretch}{1.2} 
\setlength{\tabcolsep}{4.3pt}
\begin{tabular}{|c|c|c|c|c|c|}
\hline
  \begin{tabular}[c]{@{}c@{}}Dataset\end{tabular} &
  \begin{tabular}[c]{@{}c@{}}Rule\\ Based\\ (Second)\end{tabular} &
  \begin{tabular}[c]{@{}c@{}}Supervised \\ ML \\ (Second)\end{tabular} &
  \begin{tabular}[c]{@{}c@{}}DQN-Based\\  Approach \\ (Second)\end{tabular} &
  \begin{tabular}[c]{@{}c@{}}Efficiency \\ Gain Over \\ Rule-Based\end{tabular} &
  \begin{tabular}[c]{@{}c@{}}Efficiency \\ Gain Over \\ Supervised\\  ML\end{tabular} \\ \hline
FB15k &
  60 &
  50 &
  40 &
  33.3\% &
  20\% \\ \hline
WN18 &
  65 &
  55 &
  43 &
  33.8\% &
  21.8\% \\ \hline
\end{tabular}
\caption{Efficiency in Context Integration}
\end{table}

\begin{table}[h]
\renewcommand{\arraystretch}{1.2} 
\setlength{\tabcolsep}{2.2pt}
\begin{tabular}{|c|c|c|c|c|c|c|}
\hline
Dataset &
  Metric &
  \begin{tabular}[c]{@{}c@{}}Rule\\ Based\end{tabular} &
  \begin{tabular}[c]{@{}c@{}}Supervised\\ ML\end{tabular} &
  \begin{tabular}[c]{@{}c@{}}DQN \\Based \\ Approach\end{tabular} &
  \begin{tabular}[c]{@{}c@{}}Improvement \\ Over \\ Rule \\Based (\%)\end{tabular} &
  \begin{tabular}[c]{@{}c@{}}Improvement \\ Over \\ Supervised \\ ML (\%)\end{tabular} \\ \hline
FB15k &
  \begin{tabular}[c]{@{}c@{}}KG \\ Quality \\ Index\end{tabular} &
  0.7 &
  0.75 &
  0.9 &
  28.6 &
  20 \\ \hline
WN18 &
  \begin{tabular}[c]{@{}c@{}}KG \\ Quality \\ Index\end{tabular} &
  0.68 &
  0.73 &
  0.88 &
  29.4 &
  20.5 \\ \hline
\end{tabular}
\caption{Knowledge Graph Quality Improvement}
\end{table}

\begin{figure}[htbp]
\centerline{\includegraphics[height=6cm]{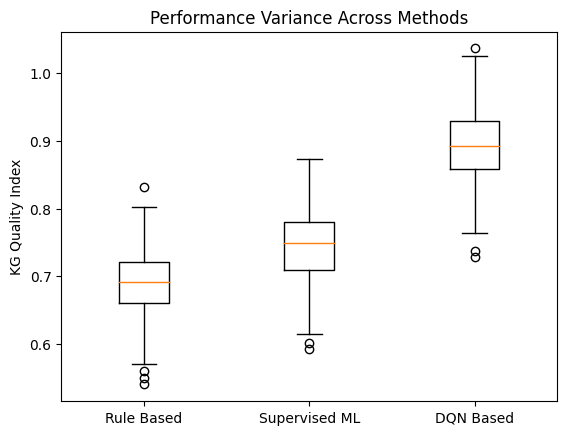}}
\caption{preformance Variance}
\label{fig-1}
\end{figure}

\begin{figure}[htbp]
\centerline{\includegraphics[height=6cm]{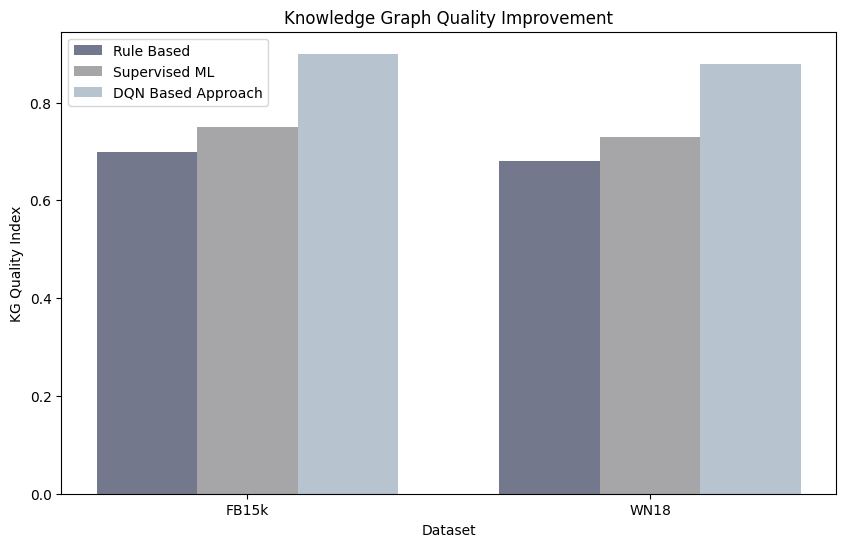}}
\caption{Knowledge Graph Quality Improvement}
\label{fig-2}
\end{figure}

\section*{Discussion}
Our research delved into using Deep Q Networks (DQN) to incorporate condensed contexts into knowledge graphs (KGs) aiming to enhance the dynamism and usefulness of KGs in fields. The inspiration, for this study stemmed from the shortcomings of supervised learning approaches, which frequently face challenges in adapting and scaling to keep up with the changing landscape of knowledge graphs.

We used a method that involved reinforcement learning focusing on using DQN to automate how context is integrated. We treated the KG as an environment. Viewed the integration actions, as choices made by an agent. Our strategy was driven by policies aimed at enhancing the quality of the KG after integration. We assessed this based on integration accuracy, efficiency and the resulting KG quality with FB15k and WN18 datasets serving as benchmarks, for our tests.

Our experiments showed how effective our DQN based method can be. We saw improvements, in how we integrated compared to using rules or supervised learning models. Our approach proved to be really good at identifying and carrying out the integration strategies. Also the time saved in integration and the better quality of our knowledge graphs show just how useful and efficient it is to use a reinforcement learning setup, for this job.

Our research adds to the increasing body of literature on using reinforcement learning in data science concerning managing knowledge graphs. The enhancements shown in the precision, speed and general quality of KGs after implementation not confirm the success of our method. Also make a strong argument, for wider use of RL methods in handling evolving data structures such, as KGs.
\section{Conclusion}
Nevertheless our study has its constraints. The intricate nature of knowledge graphs and the various methods through which contexts can be condensed require investigation, into advanced RL models and ways of integration. Moreover the ability of our method to scale up to intricate KGs and its suitability, for diverse knowledge domains are aspects that call for further examination.

In summary this research lays a groundwork, for leveraging reinforcement learning, DQN to improve the incorporation of condensed contexts into knowledge graphs. Looking ahead the ongoing development of RL models and approaches is set to pave the way for exploring opportunities in research and practical use advancing the functionalities and utility of knowledge graphs, in our data society.

\bibliographystyle{ieeetr}
\bibliography{xinde}
\end{document}